\documentclass{IOS-Book-Article}

\usepackage[utf8]{inputenc}
\usepackage[T1]{fontenc}
\usepackage{mathptmx}
\usepackage{xurl}
\usepackage{array}
\usepackage{booktabs}
\usepackage{graphicx}
\usepackage{amsmath}
\usepackage{xcolor}
\usepackage{tikz}
\usepackage{multirow}
\usetikzlibrary{arrows.meta,positioning}
\usepackage{pgfplots}

\newif\ifarxiv
\arxivtrue
\ifarxiv
\usepackage{fancyhdr}
\fancypagestyle{arxivnotice}{%
  \fancyhf{}%
  \fancyfoot[C]{\parbox{\textwidth}{\scriptsize
    \copyright\ 2026 The authors. This is the authors' version of the work. The definitive, peer reviewed and edited version will be published in \emph{Legal Knowledge and Information Systems: JURIX 2026} (Frontiers in Artificial Intelligence and Applications), IOS Press, under a CC BY-NC 4.0 license.}}%
}
\fi

\pgfplotsset{compat=1.16}

\definecolor{linknavy}{RGB}{31,73,125}
\definecolor{crejection}{HTML}{CFDCEA}
\definecolor{ctotal}{HTML}{1F497D}
\definecolor{cpartial}{HTML}{5B8DB8}
\definecolor{cannot}{HTML}{404040}
\newcommand{\cassbracket}[5]{%
  \draw[cannot, line width=0.5pt]
    ([xshift=12pt] axis cs:#1,#2) -- ([xshift=14.5pt] axis cs:#1,#2)
    -- ([xshift=14.5pt] axis cs:#1,100) -- ([xshift=12pt] axis cs:#1,100);
  \node[font=\tiny, text=cannot, rotate=90, anchor=south, inner sep=1.5pt]
    at ([xshift=14.5pt] axis cs:#1,#3) {#5};
  \node[font=\scriptsize, text=cannot, anchor=west, inner sep=1.5pt]
    at ([xshift=16pt] axis cs:#1,#3) {\textbf{#4}};
}

\definecolor{tolwine}{HTML}{882255}
\definecolor{tolrose}{HTML}{CC6677}
\definecolor{tolindigo}{HTML}{332288}
\definecolor{tolcyan}{HTML}{88CCEE}
\definecolor{tolgreen}{HTML}{117733}
\definecolor{tolteal}{HTML}{44AA99}
\definecolor{tololive}{HTML}{999933}

\newcommand{\role}[1]{%
  \tikz[baseline=(r.base)]{\node[inner xsep=2.5pt, inner ysep=0.7pt,
    rounded corners=1.5pt, fill=linknavy, text=white,
    font=\scriptsize\bfseries] (r) {#1};}}

\definecolor{hlvisa}{HTML}{FFE08A}
\definecolor{hlfam}{HTML}{B9E2F5}
\definecolor{tagvisa}{HTML}{8A5A00}
\definecolor{tagfam}{HTML}{0B5E8A}
\newcommand{\hl}[2]{{\setlength{\fboxsep}{0.7pt}\colorbox{#1}{#2}}}
\newcommand{\seglabel}[2]{\,{\tiny\upshape\bfseries\textcolor{#1}{#2}}}

\begin{document}
\begin{frontmatter}

\ifarxiv\thispagestyle{arxivnotice}\fi

\title{A Grammar of French Civil Cassation: \\ A Ground-Level Dataset and Tool}
\runningtitle{A Grammar of French Civil Cassation}

\author[A]{\fnms{Avrile} \snm{Floro}%
\thanks{Corresponding Author: Avrile Floro; E-mail: avrile.floro@ip-paris.fr.}},
\author[B]{\fnms{Tamara} \snm{Dhorasoo}},
\author[A]{\fnms{Nils} \snm{Holzenberger}},
\author[A]{\fnms{Thomas} \snm{Le Goff}},
\author[A]{\fnms{Tiphaine} \snm{Viard}}
and
\author[A]{\fnms{Maria} \snm{Boritchev}}

\runningauthor{A. Floro et al.}
\address[A]{Télécom Paris, Institut Polytechnique de Paris, France}
\address[B]{Université Polytechnique Hauts-de-France, France}

\begin{abstract}
This paper presents a dataset of 121,536 ground--reasons pairs across 86,464 rulings of the five civil chambers of the French Court of Cassation (2016--2025). An appeal on points of law cannot itself be appealed and often decides a case for good. Understanding how the Court rules is thus crucial and can help lawyers who draft those appeals. Inside a ruling, every ground of appeal (a distinct legal complaint against the appealed decision) has its own answer from the Court. This answer is called the \textit{reasons} and it belongs to one of a few \textit{doctrinal families}. Using a deterministic pipeline, we pair each ground with its reasons and attach the pair to its family, according to the established French taxonomy. We obtain a descriptive grammar of civil cassation that shows which grounds succeed, which legal texts they refer to, and how both evolve over the decade. We release the dataset, the code and an interactive consultation tool\footnotemark.
\end{abstract}

\end{frontmatter}

\footnotetext{Dataset: \href{https://doi.org/10.5281/zenodo.21932747}{DOI 10.5281/zenodo.21932747}.
Code: \href{https://github.com/avrilemay/cassation-grammar}{github.com/avrilemay/cassation-grammar}.
Consultation tool: \href{https://grammaire-cassation.fr}{grammaire-cassation.fr}. The source decisions are open data.}

\section{Introduction}\label{sec:intro}

The French Court of Cassation (\textit{Cour de cassation}) sits at the top of the judiciary, for civil and criminal matters alike. When ruling on an appeal on points of law, it can either reject it or quash the appealed decision. Its ruling is the last recourse possible and, as such, it cannot be appealed. The most common outcome is rejection ($67.5\%$ of its civil rulings over 2016--2025). Since late 2021, under the open-data mandate of France's 2016 Digital Republic Act, Judilibre\footnote{\href{https://www.courdecassation.fr/acces-rapide-judilibre}{courdecassation.fr/acces-rapide-judilibre}} (the Court's publication database) gives access to its full decisions. Publication alone, however, does not deliver knowledge of case-law trends: without structure, the relevant information is buried in the mass \cite{cadiet2018opendata}. This blind spot matters since legal scholarship documents increasingly complex laws and civil litigation~\cite{conseiletat2006securite,igj2021dossiers}. For the specialized bar that drafts the appeals on points of law, choosing the grounds to raise \emph{is} the strategy (\S\ref{sec:background}). Our objective is to turn the civil cassation corpus into a structured object that is doctrinally faithful and that can be queried. Such an instrument can inform the choice of grounds and, in the process, show how the Court rules.

A ruling contains several grounds of appeal, \textit{i.e.} the legal complaints the losing party brings against the appealed decision (\S\ref{sec:background}). The Court answers each one separately, in its reasons (\emph{motifs}, the justification it states). Each answer has its own outcome and an implicit doctrinal family: a legal category long established in French doctrine, which must be inferred from the wording of the ruling.

\paragraph{Contributions} In this paper, we contribute a dataset, extracted from Judilibre, as well as a reproducible pipeline to analyze this dataset quantitatively, in the categories of French legal doctrine. (1)~We provide $121{,}536$ ground--reasons pairs covering the five civil chambers of the Court of Cassation from 2016 to 2025. The 117,073 pairs whose ground is accepted or rejected are attached to their doctrinal family. An article-level layer records the legal texts the Court relies on. (2)~We release the deterministic rule-based method that derives these labels, audited at every stage by legally trained annotators. (3)~From these counts, we draw a first descriptive grammar of civil cassation accessible through an interactive tool, as an example of the analyses the dataset makes possible.

\section{Related Work}\label{sec:related}

\paragraph{Predicting the outcome of a case} The most common way to study court decisions computationally is through legal judgment prediction: giving a model the text of a case and letting it predict the outcome~\cite{chalkidis2019neural,katz2017general,xiao2018cail}, pioneered on the European Court of Human Rights (ECtHR)~\cite{aletras2016predicting}.
The work closest to ours is Sulea et al.~\cite{sulea2017predicting}, who classify decisions of the Court of Cassation and report strong performance at predicting the case-level ruling. A ruling of the Court often rejects some grounds and quashes others inside the same decision (\S\ref{sec:res_mixed}), which a single label per case cannot show. We therefore lower the unit of analysis from the ruling to the single ground. For French domestic case law, the existing resources stay at the level of the whole decision: linked first-instance and appellate judgments~\cite{condevaux2022linking} and outcome-labeled court of appeal rulings on a single issue~\cite{vaudaux-etal-2023-pretrained}. To our knowledge, no ground-level resource exists for the Court of Cassation.

\paragraph{Recovering the internal structure of a decision} A different line of work tries to recover what happens inside a judgment instead of only its final outcome. Rhetorical-role labeling tags each sentence with what it does in the text, for instance in Indian Supreme Court judgments~\cite{Saravanan_10.1007/s10506-010-9087-7,bhattacharya2019identification,malik2022semantic}. In our work, rather than segmenting the decision from scratch, we build on the zoning that Judilibre already provides: the division of each ruling into functional zones (\S\ref{sec:corpus}). We then link each argument to the passage that responds to it (\S\ref{sec:pairing}). This pairing of a ground with the Court's reasons is, in practice, a relational form of argument mining, which detects, types and links the arguments of a decision~\cite{mochales2011argumentation}, recently revisited with LLMs and few-shot methods~\cite{xu2025labelaware}. Related annotation efforts structure civil-law and EU decisions semantically or by interpretative function~\cite{zorzanellicosta2024annotation,grundler2025jif}, or annotate ECtHR citations in context and release an exploration tool~\cite{mumford2025context}.

\paragraph{Rule-based methods and interpretability} Explicit rules are the oldest way of treating legal text by computer~\cite{sergot_1986_10.1145/5689.5920} and did not disappear when machine learning arrived~\cite{chiticariu-etal-2013-rule}. They remain a preferred first pass on clearly patterned legal text~\cite{nan2024combining}, including for segmenting court decisions from open legal data~\cite{darji2025segmentation}. Rudin~\cite{rudin2019stop} argues that high-stakes settings call for 
interpretable models rather than explained black boxes. Expert-curated factor lists still outperform LLM-discovered ones for case outcome prediction~\cite{gray2024using}. In our system, each outcome points back to a doctrinal family drawn from Boré's taxonomy~\cite{bore2023cassation} (\S\ref{sec:background}): a jurist can retrace every decision and disagree on legal terms. 

\paragraph{French legal NLP and language models} The reference language model is JuriBERT~\cite{douka2021juribert}, a BERT model pre-trained on legal French including pseudonymized Court of Cassation decisions~\cite{girardchanudet2023justice}, which is the data we work on. On the same court, Charmet et al.~\cite{charmet-etal-2022-complex} train models to generate the keyword sequences that index the rulings. A final group of studies measures how far one can trust a large language model on legal tasks~\cite{dahl2024large,guha2023legalbench}. We build the resource itself with deterministic rules. We use fine-tuned models, JuriBERT among them, as baselines against those rules (\S\ref{sec:baseline}).

\section{Background: How the Court of Cassation Writes}\label{sec:background}

\paragraph{A court of law, not of facts} The Court of Cassation is not a third instance since it does not re-try cases nor re-examine the evidence: it reviews whether the law was properly applied by the lower court. A party who has lost before a court of appeal files an appeal on points of law. If the court of appeal erred in law, the decision is quashed by the Court of Cassation and remitted to a different one. A cassation can be total (the whole appealed decision falls) or partial (only part of it is quashed).

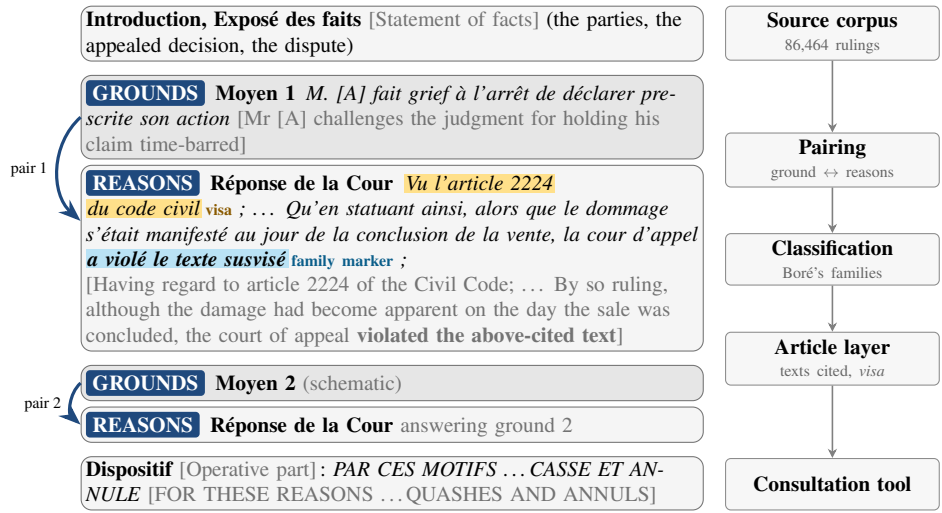
\begin{figure}[t]
\centering
\hspace*{-3mm}
\begin{tikzpicture}[
  font=\scriptsize,
  zone/.style={draw=black!55, rounded corners, align=left, inner sep=1.8pt, text width=8.1cm,
               minimum height=0.34cm},
  stg/.style={draw=black!45, rounded corners=2pt, align=center, inner sep=2.5pt, text width=2.65cm,
              minimum height=0.68cm, fill=black!3},
  >=Stealth]
  \node[zone, fill=black!4] (intro)
    {\textbf{Introduction, Exposé des faits} \textcolor{black!55}{[Statement of facts]} (the parties, the appealed decision, the dispute)};
  \node[zone, fill=black!10, below=5pt of intro] (m1)
    {\role{GROUNDS}\enspace\textbf{Moyen 1}\enspace \emph{M. [A] fait grief à l'arrêt de déclarer
     prescrite son action} \textcolor{black!55}{[Mr [A] challenges the judgment for holding his
     claim time-barred]}};
  \node[zone, fill=black!4, below=2pt of m1] (r1)
    {\role{REASONS}\enspace\textbf{Réponse de la Cour}\enspace \emph{\hl{hlvisa}{Vu l'article 2224} \hl{hlvisa}{du code civil}\seglabel{tagvisa}{visa} ; \dots\ Qu'en statuant ainsi, alors que le dommage s'était manifesté au jour de la
     conclusion de la vente, la cour d'appel \hl{hlfam}{\textbf{a violé le texte susvisé}}\seglabel{tagfam}{family~marker} ;}\\
     \textcolor{black!55}{[Having regard to article 2224 of the Civil Code; \dots\ By so ruling,
     although the damage had become apparent on the day the sale was concluded, the court of
     appeal \textbf{violated the above-cited text}]}};
  \node[zone, fill=black!10, below=5pt of r1] (m2)
    {\role{GROUNDS}\enspace\textbf{Moyen 2} \textcolor{black!55}{(schematic)}};
  \node[zone, fill=black!4, below=2pt of m2] (r2)
    {\role{REASONS}\enspace\textbf{Réponse de la Cour} \textcolor{black!55}{answering ground 2}};
  \node[zone, fill=black!4, below=5pt of r2] (disp)
    {\textbf{Dispositif} \textcolor{black!55}{[Operative part]}\,: \emph{PAR CES MOTIFS \dots CASSE ET ANNULE}
    \textcolor{black!55}{[FOR THESE REASONS \dots QUASHES AND ANNULS]}};
  \draw[->, very thick, linknavy] (m1.west) to[bend right=48] node[left, font=\tiny, black]{pair 1} ([yshift=14pt]r1.west);
  \draw[->, very thick, linknavy] (m2.west) to[bend right=48] node[left, font=\tiny, black]{pair 2} (r2.west);
  \node[stg, anchor=north west] at ([xshift=8pt]intro.north east) (s1)
    {\textbf{Source corpus}\\[-1pt] {\tiny\color{black!60}86,464 rulings}};
  \node[stg, anchor=south west] at ([xshift=8pt]disp.south east) (s5)
    {\textbf{Consultation tool}};
  \path (s1.south) -- (s5.north) coordinate[pos=0.25] (p2) coordinate[pos=0.5] (p3) coordinate[pos=0.75] (p4);
  \node[stg, anchor=center] at (p2 -| s1) (s2)
    {\textbf{Pairing}\\[-1pt] {\tiny\color{black!60}ground $\leftrightarrow$ reasons}};
  \node[stg, anchor=center] at (p3 -| s1) (s3)
    {\textbf{Classification}\\[-1pt] {\tiny\color{black!60}Boré's families}};
  \node[stg, anchor=center] at (p4 -| s1) (s4)
    {\textbf{Article layer}\\[-1pt] {\tiny\color{black!60}texts cited, \emph{visa}}};
  \foreach \a/\b in {s1/s2, s2/s3, s3/s4, s4/s5}
    \draw[-{Stealth[length=3.5pt, width=3.5pt]}, black!50, line width=0.6pt] (\a.south) -- (\b.north);
\end{tikzpicture}
\caption{The zones of a ruling (left, adapted from Civ.\,3, 2019, no.\,18-12.682) and the pipeline that reads them (right). Only the badged zones are extracted. Highlighted: the matched \emph{visa} and family marker.}
\label{fig:zones}
\end{figure}

\paragraph{Anatomy of a ruling} Rulings follow a near-fixed sequence of functional zones (Figure~\ref{fig:zones}). In October 2019, the Court adopted a direct-style drafting \cite{cassation_guide_redaction}. Now, the reasons start with « Réponse de la Cour » [The Court's reasons], while before they opened with « Attendu que » [Whereas]. When a ground succeeds, the reasoning usually starts with a \emph{visa} (« Vu l'article \dots » [Having regard to article \dots]). This is the legal footing which lists the texts the Court relies on. Grounds and reasons alternate down the ruling.

\paragraph{How an appeal on points of law is built}\label{para:appeal_built} Appeals on points of law are drafted by a specialized bar, the \emph{avocats aux Conseils}, who hold a monopoly on representation before the Court of Cassation (Code of Civil Procedure, hereafter CPC, art.\,973). They were 132 in 2025 \cite{adlc2025avis}. They file a statement of grounds of appeal 
in which their legal criticisms are presented as a numbered list of grounds (CPC, art.\,978). A ground attacks one or several heads of the appealed decision (the precise points the court of appeal decided). Grounds can split further into branches: numbered sub-arguments, a point of law each.

\paragraph{The families of cassation\label{sec:bg_cassation}}

When the Court quashes, doctrine sorts the why into a small set of families. The taxonomy was settled long ago and its canonical modern statement is Boré's treatise, continuously re-edited since 1980~\cite{bore2023cassation} and written by two presidents of the \textit{Ordre des avocats aux Conseils}. The list is deliberately not statutory. Consulted on the draft Code of Civil Procedure, the \textit{Conseil d'État} (France's supreme administrative court) advised against an exhaustive enumeration, so that the Court stays, as Boré puts it, master of its own openings~\cite{bore2023cassation}. The criminal openings, by contrast, are fixed by the Code of Criminal Procedure~\cite{bore2024penale}. Table~\ref{tab:bg_cassation} lists the~families as they stand today.

\begin{table}[t]
\centering
\caption{The doctrinal families of cassation.}
\label{tab:bg_cassation}
\scriptsize
\setlength{\tabcolsep}{3pt}
\begin{tabular}{>{\raggedright\arraybackslash}p{2.2cm}p{4.0cm}p{5.5cm}}
\toprule
Doctrinal family & Description & Typical wording \\
\midrule
Violation of the law & A legal text was applied or interpreted incorrectly. & « \emph{la cour d'appel a violé le texte susvisé}~» [the court of appeal violated the above-cited text] \\
Lack of legal basis & The factual findings are incomplete, so the Court cannot check that the law was correctly applied. & « \emph{sans rechercher si \dots, la cour d'appel a privé sa décision de base légale}~» [without investigating whether \dots, the court deprived its decision of legal basis] \\
Distortion & The court misread a clear and precise document: it betrayed its plain meaning. & « \emph{en dénaturant les termes clairs et précis de~\dots}~» [by distorting the clear and precise terms of \dots] \\
Defect of reasons & The reasoning is absent, contradictory, doubtful or fails to answer a party's submissions. & « \emph{sans répondre aux conclusions \dots, la cour d'appel n'a pas satisfait aux exigences du texte susvisé}~» [without answering the submissions
 \dots, the court did not meet the requirements of the above-cited text] \\
Excess of power & The court overstepped its jurisdiction or its office. & « \emph{la cour d'appel a excédé ses pouvoirs}~» [the court of appeal exceeded its powers] \\
Breach of procedure & A procedural rule was violated, most often the adversarial principle. & « \emph{violation du principe de la contradiction}~» [violation of the adversarial principle] \\
Omission / \emph{ultra petita} & The court failed to rule on a claim, or ruled beyond what was claimed. & « \emph{en statuant sur des choses non demandées}~» [by ruling on matters not claimed] \\
Contrariety of judgments & Two irreconcilable decisions were rendered between the same parties. & « \emph{deux décisions inconciliables}~» [two irreconcilable decisions] \\
\bottomrule
\end{tabular}
\end{table}

\paragraph{The families of rejection\label{sec:bg_rejection}}

Rejection has its families too: doctrine~\cite{bore2023cassation} distinguishes the three situations of Table~\ref{tab:bg_rejection}. Boré treats an inoperative ground as a form of inadmissibility, and we report the two together as one family.

\begin{table}[t]
\centering
\caption{The rejection families.}
\label{tab:bg_rejection}
\scriptsize
\setlength{\tabcolsep}{3pt}
\begin{tabular}{>{\raggedright\arraybackslash}p{2.2cm}p{5.5cm}p{4.0cm}}
\toprule
Rejection family & Description & Typical wording \\
\midrule
Summary rejection & The Court just declares that the ground cannot lead to cassation, and does not say why. & « \emph{pas lieu de statuer par une décision spécialement motivée}~» [no specially reasoned decision needed] \\
On the merits & The reasons discuss the ground's legal substance and find it wanting. & « \emph{le moyen n'est pas fondé}~» [the ground is ill-founded] \\
\multirow[t]{2}{=}{Inadmissibility and inoperativeness (one family, following Boré)} & Inadmissible: the ground cannot even be examined, for instance because it is raised for the first time before the Court. & « \emph{le moyen est irrecevable}~» [the ground is inadmissible] \\
 & Inoperative: the ground may be right, yet cannot change the outcome. It attacks a block of reasons the decision did not rely on. & « \emph{le moyen est inopérant}~» [the ground is inoperative] \\
\bottomrule
\end{tabular}
\end{table}

\section{Corpus Construction}\label{sec:corpus}

\paragraph{Source}\label{para:corpus:source} Our corpus is made of the rulings of the Court of Cassation. We retrieved it from Judilibre (\S\ref{sec:intro}), accessed through PISTE, the public French API portal. We collected the data once, in May 2026. We retrieved $548{,}126$ decisions over 1947--2026. They contain more than the text of the ruling: procedural metadata (chamber, formation, decision date) and an outcome label (the \textit{decision field}) indicating whether the ruling is a cassation, a rejection, or something else. Whether a cassation is total or partial is not in the metadata. We derive it from the operative part (which either quashes the whole decision or lists the quashed heads). Crucially, the section offsets are also available, showing where each functional zone of the decision starts and ends (including the Court's reasons).

\paragraph{Scope} The Court of Cassation has five civil chambers and a criminal one (Code of Judicial Organization, hereafter COJ, art. R.\,421-3). We focus on the civil chambers (first, second and third civil, commercial, and social). The criminal chamber is excluded: its openings for cassation follow a different logic (\S\ref{sec:bg_cassation}) and its procedure differs substantially \cite{bore2024penale}. We only keep the rulings that decide an appeal on points of law. Hence, non-judgment decisions, petitions and withdrawals are discarded (the \textit{petition filter}).  The time period runs from 2016 to 2025. Indeed, coverage of section offsets for cassations jumps from almost none ($0$--$1\%$ in 2008--2015) to $91$--$100\%$ from 2016 on. We obtain $86{,}464$ cassation and rejection rulings. The source decisions are pseudonymized by the Court and we remove residual names before release. Statistics are aggregated by chamber, not by individual judge, as French law prohibits judicial profiling (COJ, art. L.\,111-13).

\section{Method}\label{sec:method}

One pair is one ground and the block of the Court's reasons that answers it. Branches are not counted separately. While the ground is the unit we count, the Court's reasons are the only text our rules read (Figure~\ref{fig:zones}). 

\subsection{Pairing Each Ground with the Court's Reasons}\label{sec:pairing}

The alternating construction of the ruling (\S\ref{sec:background}) makes the pairing natural. We first align grounds to reasons by position: the $i$-th ground pairs with the $i$-th block of reasons. Equal counts on both sides indicate that nothing was split or lost. 
They differ in $14{,}826$ rulings. We use deterministic rules to re-merge split grounds 
or blocks of reasons and to give lone blocks of reasons a pair of their own. The rules use the drafting conventions, such as ground numbering or opening and closing formulas. After this stage, we reach $121{,}536$ pairs over $86{,}464$ rulings. Roughly $32\%$ of pairs carry no ground text anywhere in their ruling. Those are almost all rejections, since in a summary rejection the ground is often sent to an annex instead of being reproduced in the main ruling. In those cases, the pipeline builds a degenerate pair (an empty ground facing a block of reasons), which does not impact classification since only the reasons are read. 
A few cassations ($\approx 1.3\%$ of the accepted grounds) also come without their ground text, due to joined appeals, grounds the Court raised on its own, or zoning gaps.

\subsection{Classifying the Outcome of Each Ground}\label{sec:outcome}

Every pair receives one of the six outcomes of Table~\ref{tab:statuts}. Three statuses are linked to the legal substance: accepted (the ground succeeds), rejected (the ground fails), and consequential cassation (a head falls with the head it depended on). The other three are structural only: no-match, out-of-scope and conflict. The status is derived from a fixed list of regular-expression detectors. They first scan the reasons for cassation signals, then run the same text through the separate rejection pipeline. If the reasons match both pipelines and belong to a rejection ruling, the pair is classified as rejected, since no ground succeeds in such a ruling. When the ruling is a cassation (partial or total), we do not force a classification. As a result, $155$ pairs keep the conflict status. When the reasons match neither pipeline, two filters apply: one for the wording of a consequential cassation, one for the closing ``Scope and consequences of the cassation'' section, which answers no ground and is treated as out-of-scope. What remains is classified as no-match.

\begin{table}[t]
\centering
\caption{The six outcome statuses and their distribution over the final corpus.}
\label{tab:statuts}
\scriptsize
\setlength{\tabcolsep}{4pt}
\begin{tabular}{llrr}
\toprule
Status & Meaning & Pairs & Share \\
\midrule
Rejected      & the ground fails                    & 84{,}460 & $69.5\%$ \\
Accepted     & the ground succeeds, cassation         & 32{,}613 & $26.8\%$ \\
No-match & no signal found & 2{,}735 & $2.3\%$ \\
Consequential                  & cassation by consequence            & 907 & $0.7\%$ \\
Out-of-scope & not a ground on the merits         & 666 & $0.5\%$ \\
Conflict     & contradictory signals              & 155 & $0.1\%$ \\
\bottomrule
\end{tabular}
\end{table}

\subsection{Doctrinal Families: How the Court Quashes, Why It Rejects}\label{sec:grounds}

The outcome says who wins, which Judilibre already gives at the ruling level (\S\ref{sec:corpus}). The doctrinal family says why the Court quashed or refused to quash, which is novel. For the accepted grounds, we follow the doctrinal families of \S\ref{sec:bg_cassation}: we read the Court's reasons with a grid of 149 regular-expression detectors, applied as a cascade. We start from the most specific verbal marker down to the most residual one (violation of the law). 
On the rejection side, a second grid of 110 patterns sorts each ground into the three families of~\S\ref{sec:bg_rejection}. 
This assignment is multi-label: a ground can, for instance, be inadmissible in one branch and ill-founded in another. Both grids were written from the canonical formulas of each family, as found in doctrine and in Boré's treatise~\cite{bore2023cassation}, then refined on the blocks they left unmatched, before the gold annotation.
These labels form a silver set (automatically generated). Section~\ref{sec:validation} validates them against a hand-annotated gold set.

\subsection{From Grounds to Texts: Article-Level Granularity}\label{sec:articles}

A final stage scans each block of reasons for references to articles and codes. Each reference gives the article's code and number, whether it is cited in the \emph{visa} (\S\ref{sec:background}) or only in the body of the reasons. Each reference inherits its pair's outcome: an article is \textit{accepted} when its ground is. The 2016 reform of the law of obligations renumbered the Civil Code. Where possible, we map the old numbering onto the new (e.g. article~1382\,$\leftrightarrow$\,1240).

\section{Validation}\label{sec:validation}


\paragraph{Annotation protocol} All annotation is blind to the machine labels and performed by annotators with at least five years of legal education each. The structural stages (decision field, petition filter, ground pairing, unusual statuses) are audited by a single annotator. The doctrinal families are doubly annotated: two annotators independently read the same $400$ grounds ($200$ accepted, $200$ rejected). Their disagreements are adjudicated between them. The adjudicated gold is compared with the machine labels.

\subsection{Sampling Design}\label{sec:sampling}

Sample sizes are in Table~\ref{tab:validation}. Two stages are drawn uniformly, for unbiased estimates: (1)~the decision field, drawn before the scope filter and relabeled outcome by outcome, and (2) the pairing, where each ground is checked against the reasons answering it. Targeted draws focus on the error-prone cases: (1) the petition filter (\S\ref{sec:corpus}) is audited on decisions worded like petitions, and (2) six pools of 100 cover the four unusual statuses of Table~\ref{tab:statuts} (no-match, conflict, out-of-scope, consequential) and the blocks with no reproduced ground (accepted and summary-rejection, \S\ref{sec:pairing}), plus 20 uniform controls. The family campaign mixes the two regimes: on each side, $50$ grounds are drawn uniformly and $150$ are stratified, so that rare families are covered. Aggregate figures are then reweighted to the corpus distribution.

\subsection{Results by Stage}\label{sec:valresults}

Table~\ref{tab:validation} gathers the figures. Structural stages are reported as percent agreement with the machine label. The pairing audit finds a single pair merging two grounds. Among the unusual statuses, precision goes from 95\% (out-of-scope) to 100\% where a machine label can be confirmed. The no-match pool is roughly half non-merits material, half real misses of the rules. Table~\ref{tab:iaa} reports the double annotation of the doctrinal families, measured with Cohen's $\kappa$~\cite{cohen1960kappa}. On the sample, per-family $\kappa$ ranges from 0.90 to 0.99 (two classes are too rare to measure). After reweighting, all families stay at $\kappa \geq 0.96$ except inadmissibility/inoperativeness, which drops to 0.83: it represents only 6.2\% of rejections (\S\ref{sec:res_reject}). $\kappa$ corrects for chance more severely on a rare class, and 9 of the 10 rejection disagreements involve this label. All 17 (10 rejected, 7 accepted) were adjudicated jointly into a gold set, which confirms the machine label on 90.5\% of accepted grounds (181/200) and 91.5\% of rejected ones (183/200). Table~\ref{tab:baseline} scores all baselines against it.

\begin{table}[t]
\begin{minipage}[t]{0.56\linewidth}
\centering
\tablewidth=\hsize
\caption{Validation by stage.}
\label{tab:validation}
\scriptsize
\setlength{\tabcolsep}{3pt}
\begin{tabular}{@{}llr@{}}
\toprule
Stage & Sampling ($n$) & Agreement \\
\midrule
Decision field      & uniform ($200$)  & 99.5\% \\ 
Petition filter     & targeted ($200$) & 99.0\% \\
Ground pairing      & uniform ($200$)  & 99.5\% \\
Unusual statuses    & targeted ($620$) & $95$--$100\%$ \\
Doctrinal families  & double annot.\ ($2\times200$) & Table~\ref{tab:iaa}\\
\bottomrule
\end{tabular}
\end{minipage}\hfill
\begin{minipage}[t]{0.41\linewidth}
\centering
\tablewidth=\hsize
\caption{Agreement of the baselines with the held-out gold ($197$ accepted, $200$ rejected).}
\label{tab:baseline}
\scriptsize
\setlength{\tabcolsep}{4pt}
\begin{tabular}{@{}lcc@{}}
\toprule
 & Accepted & Rejected \\
\midrule
\textbf{Rules (this paper)} & \textbf{91.4\%} & \textbf{91.5\%} \\
JuriBERT           & $89.3\%$ & $92.0\%$ \\
CamemBERTv2        & $78.7\%$ & $93.5\%$ \\
Qwen3-32B (0 shot) & $87.3\%$ & $88.5\%$ \\
\bottomrule
\end{tabular}
\end{minipage}
\end{table}

\begin{table}[t] 
\centering 
\tablewidth=\hsize
\caption{Double annotation of the doctrinal families.} 
\label{tab:iaa} 
\scriptsize 
\setlength{\tabcolsep}{3pt} 
\begin{tabular}{@{}llcccc@{}} 
\toprule 
& & \multicolumn{2}{c}{Sample} & \multicolumn{2}{c}{Corpus (reweighted)} \\
\cmidrule(lr){3-4}\cmidrule(l){5-6} 
Outcome ($n$) & Reading & $\kappa$ [$95\%$ CI] & Agreement & $\kappa$ [$95\%$ CI] & Agreement \\ 
\midrule 
Accepted ($200$) & single-label & $0.957$ [$0.922$--$0.987$] & $96.5\%$ & $0.961$ [$0.917$--$0.993$] & $97.5\%$ \\ 
Rejected ($200$) & exact sets & $0.932$ [$0.889$--$0.972$] & $95.0\%$ & $0.929$ [$0.876$--$0.973$] & $95.9\%$ \\ 
\bottomrule 
\end{tabular} 
\end{table}

\paragraph{Baselines\label{sec:baseline}}
We want to check whether the released labels are learnable and whether a model beats the rules. We fine-tune the base versions of JuriBERT (\S\ref{sec:related}) and CamemBERTv2~\cite{antoun2024camembert2} to predict the doctrinal families from the silver labels, reading the blocks of reasons. Predictions mirror the dataset's label structure (\S\ref{sec:grounds}): single-label over the six main cassation families (Table~\ref{tab:bg_cassation}, without omission and contrariety) for accepted grounds, and multi-label over the three rejection families of Table~\ref{tab:bg_rejection} for rejected ones
(balanced training, learning rate $2\times10^{-5}$, effective batch $32$, at most four epochs, checkpoint selection on a held-out split). Evaluation uses a held-out gold set of $200$ rejected and $197$ accepted grounds: the three accepted grounds whose family lies outside the six the models predict are dropped. Both fine-tuned models beat the rules on rejected grounds and stay below them on accepted ones, the generalist model falling far behind on accepted grounds. A zero-shot LLM (Qwen3-32B \cite{qwen3technicalreport}) is given the annotators' guideline and their answer options and stays below the rules. JuriBERT errors concentrate on the cases where the gold and the rules already disagree ($82\%$ of them, against $1$--$4\%$ of the agreement cases). The zero-shot LLM shows a much weaker concentration ($18$--$35\%$ against $9$--$12\%$ of the agreement cases). Trained on silver labels, the fine-tuned models inherit their blind spots.

\section{Results: A Grammar of Civil Cassation}\label{sec:results}

\paragraph{How appeals on points of law resolve\label{sec:res_outcome}}

\begin{figure}[t]
\centering
\begin{tikzpicture}
\begin{axis}[
    ybar stacked,
    bar width=15pt,
    width=0.97\linewidth, height=5.0cm,
    ymin=0, ymax=100,
    ylabel={\scriptsize \% of rulings},
    xmin=0.5, xmax=5.9,
    xtick={1,...,5},
    xticklabels={First civil, Second civil, Third civil, Commercial, Social},
    x tick label style={font=\scriptsize},
    y tick label style={font=\scriptsize},
    ytick={0,20,...,100},
    axis line style={draw=gray!60},
    tick style={draw=gray!60},
    axis lines=left,
    legend style={
        at={(0.5,-0.18)}, anchor=north,
        legend columns=-1, draw=none, font=\scriptsize,
        /tikz/every even column/.append style={column sep=10pt},
    },
    legend cell align=left,
    clip=false,
]
\addlegendimage{area legend, fill=crejection, draw=black, line width=0.3pt}
\addlegendimage{area legend, fill=ctotal, draw=black, line width=0.3pt}
\addlegendimage{area legend, fill=cpartial, draw=black, line width=0.3pt}
\addplot[forget plot, ybar, fill=crejection, draw=none, point meta=explicit,
    nodes near coords, every node near coord/.append style={anchor=center, font=\scriptsize, text=cannot}]
    coordinates {(1,69) [69] (2,64) [64] (3,69) [69] (4,69) [69] (5,67) [67]};
\addplot[forget plot, ybar, fill=ctotal, draw=none, point meta=explicit,
    nodes near coords, every node near coord/.append style={anchor=center, font=\scriptsize, text=white}]
    coordinates {(1,13) [13] (2,21) [21] (3,12) [12] (4,13) [13] (5,8) [8]};
\addplot[forget plot, ybar, fill=cpartial, draw=none, point meta=explicit,
    nodes near coords, every node near coord/.append style={anchor=center, font=\scriptsize, text=white}]
    coordinates {(1,18) [18] (2,15) [15] (3,19) [19] (4,18) [18] (5,25) [25]};
\legend{Rejection, Total cassation, Partial cassation}
\cassbracket{1}{69}{84.5}{31}{41 / 59}
\cassbracket{2}{64}{82}{36}{58 / 42}
\cassbracket{3}{69}{84.5}{31}{39 / 61}
\cassbracket{4}{69}{84.5}{31}{43 / 57}
\cassbracket{5}{67}{83.5}{33}{24 / 76}
\end{axis}
\end{tikzpicture}
\caption{How appeals on points of law resolve, by chamber (2016--2025).}
\label{fig:outcome}
\end{figure}
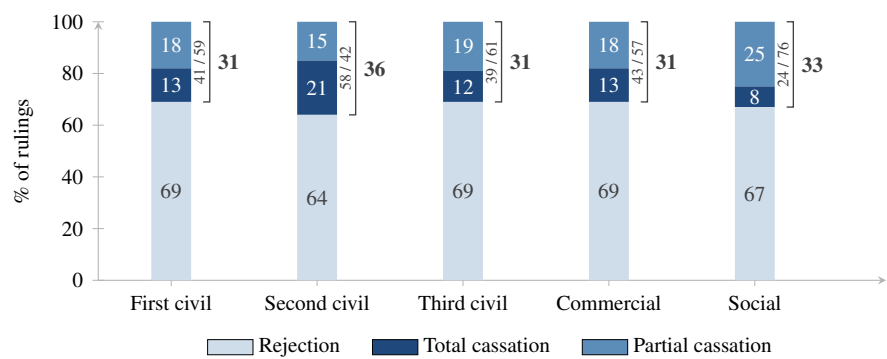

The 86,464 civil rulings of 2016--2025 split into 28,144 cassations ($32.5\%$) and 58,320 rejections. The cassations are more often partial than total ($59.8\%$ partial). The cassation rate is almost the same across the five chambers, from $31\%$ to $36\%$, but its form is not: total cassations dominate only in the second civil chamber ($58\%$), against $24\%$ in the social chamber (Figure~\ref{fig:outcome}). The Court also decides each ground on its own, so one ruling can simultaneously accept a ground and reject another\label{sec:res_mixed}. These \emph{mixed rulings} account for more than one cassation ruling in three ($38.8\%$). The total/partial divide is about the decision (how much of it falls), not the grounds. 

\paragraph{The grammar of cassation\label{sec:res_accept}}

A doctrinal family is attached to $97.9\%$ of cassation rulings. The distribution across families is highly concentrated  (Table~\ref{tab:families}). Violation of the law ($56.9\%$) and lack of legal basis ($23.2\%$) amount to $80.1\%$ of all accepted grounds. The rest is a long tail. Two families of Table~\ref{tab:bg_cassation} are absent in practice: omission (one detected pair, an \emph{ultra petita}) and contrariety of judgments (no detected pair). Omissions are repaired by the court that made them (CPC, art.\,463) and a successful contrariety ends in an annulment, outside our scope (\S\ref{sec:corpus}). The distribution of families varies by \mbox{chamber (Table~\ref{tab:famch}).}

\begin{table}[t]
\begin{minipage}[t]{0.40\linewidth}
\centering
\tablewidth=\hsize
\caption{Accepted grounds of cassation.}
\label{tab:families}
\scriptsize
\setlength{\tabcolsep}{3pt}
\begin{tabular}{@{}lrr@{}}
\toprule
Doctrinal family & Grounds & Share \\
\midrule
Violation of the law                & $18{,}542$ & $56.9\%$ \\
Lack of legal basis                 & $7{,}572$ & $23.2\%$ \\
Defect of reasons                   & $2{,}073$ & $6.4\%$ \\
Excess of power                     & $1{,}920$ & $5.9\%$ \\
Breach of procedure                 & $1{,}255$ & $3.8\%$ \\
Distortion  & $1{,}250$ & $3.8\%$ \\
Omission / \emph{ultra petita}             & $1$ & $0.0\%$ \\
\bottomrule
\end{tabular}
\end{minipage}\hfill
\begin{minipage}[t]{0.57\linewidth}
\centering
\tablewidth=\hsize
\caption{Family shares within each chamber's accepted grounds (\%).}
\label{tab:famch}
\scriptsize
\setlength{\tabcolsep}{3pt}
\begin{tabular}{@{}lrrrrr@{}}
\toprule
 & Civ.\,1 & Civ.\,2 & Civ.\,3 & Com. & Soc. \\
\midrule
Violation of the law & 54.8 & 61.7 & 52.7 & 50.4 & 60.1 \\
Lack of legal basis  & 23.3 & 19.5 & 23.2 & 29.1 & 22.8 \\
Defect of reasons    &  5.9 &  4.6 &  9.9 &  7.3 &  5.6 \\
Excess of power      &  6.4 &  5.1 &  6.1 &  4.9 &  6.7 \\
Breach of procedure  &  5.2 &  5.4 &  3.8 &  3.4 &  2.2 \\
Distortion           &  4.4 &  3.7 &  4.3 &  4.9 &  2.7 \\
\bottomrule
\end{tabular}
\end{minipage}
\vspace{-0.1cm}
\end{table}

\paragraph{The grammar of rejection\label{sec:res_reject}}

The rejection families are multi-label (\S\ref{sec:grounds}) and $99.6\%$ of the rejection rulings are associated with at least one rejection mechanism. Over the $84{,}460$ rejected grounds, the shares overlap and do not add up to one. Overall, summary rejection covers $61.9\%$ of these grounds, while $38.4\%$ of them are dismissed on the merits. Rejections for inadmissibility or inoperativeness are fairly rare, at $6.2\%$.

\paragraph{Movement over the decade\label{sec:res_time}}

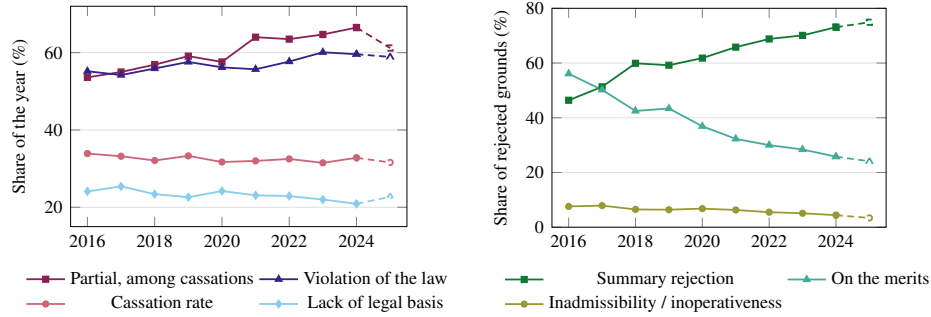
\begin{figure}[t]
\centering
\resizebox{0.485\linewidth}{!}{%
\begin{tikzpicture}
\begin{axis}[
  width=6.8cm, height=5cm,
  xmin=2015.5, xmax=2025.5, ymin=15, ymax=72,
  xtick={2016,2018,2020,2022,2024},
  x tick label style={font=\scriptsize, /pgf/number format/1000 sep=},
  y tick label style={font=\scriptsize},
  ylabel={\scriptsize Share of the year (\%)},
  ymajorgrids, grid style={black!12},
  legend style={font=\scriptsize, at={(0.5,-0.16)}, anchor=north, legend columns=2, draw=none},
  every axis plot/.append style={thick},
]
\addplot[color=tolwine, mark=square*, mark size=1.2pt] coordinates {(2016,53.6)(2017,55.0)(2018,56.9)(2019,59.1)(2020,57.6)(2021,64.0)(2022,63.5)(2023,64.7)(2024,66.5)};
\addlegendentry{Partial, among cassations}
\addplot[color=tolwine, densely dashed, mark=square, mark size=1.2pt, forget plot] coordinates {(2024,66.5)(2025,61.2)};
\addplot[color=tolindigo, mark=triangle*, mark size=1.5pt] coordinates {(2016,55.2)(2017,54.2)(2018,55.9)(2019,57.6)(2020,56.2)(2021,55.7)(2022,57.7)(2023,60.1)(2024,59.6)};
\addlegendentry{Violation of the law}
\addplot[color=tolindigo, densely dashed, mark=triangle, mark size=1.5pt, forget plot] coordinates {(2024,59.6)(2025,58.9)};
\addplot[color=tolrose, mark=*, mark size=1.2pt] coordinates {(2016,33.9)(2017,33.2)(2018,32.1)(2019,33.3)(2020,31.7)(2021,32.0)(2022,32.5)(2023,31.5)(2024,32.8)};
\addlegendentry{Cassation rate}
\addplot[color=tolrose, densely dashed, mark=o, mark size=1.2pt, forget plot] coordinates {(2024,32.8)(2025,31.6)};
\addplot[color=tolcyan, mark=diamond*, mark size=1.5pt] coordinates {(2016,24.1)(2017,25.4)(2018,23.4)(2019,22.6)(2020,24.2)(2021,23.1)(2022,22.9)(2023,22.0)(2024,20.9)};
\addlegendentry{Lack of legal basis}
\addplot[color=tolcyan, densely dashed, mark=diamond, mark size=1.5pt, forget plot] coordinates {(2024,20.9)(2025,22.7)};
\end{axis}
\end{tikzpicture}}%
\hfill
\resizebox{0.485\linewidth}{!}{%
\begin{tikzpicture}
\begin{axis}[
  width=6.8cm, height=5cm,
  xmin=2015.5, xmax=2025.5, ymin=0, ymax=80,
  xtick={2016,2018,2020,2022,2024},
  x tick label style={font=\scriptsize, /pgf/number format/1000 sep=},
  y tick label style={font=\scriptsize},
  ylabel={\scriptsize Share of rejected grounds (\%)},
  ymajorgrids, grid style={black!12},
  legend style={font=\scriptsize, at={(0.5,-0.16)}, anchor=north, legend columns=2, draw=none},
  every axis plot/.append style={thick},
]
\addplot[color=tolgreen, mark=square*, mark size=1.2pt] coordinates {(2016,46.4)(2017,51.3)(2018,59.9)(2019,59.2)(2020,61.8)(2021,65.8)(2022,68.8)(2023,70.1)(2024,73.1)};
\addlegendentry{Summary rejection}
\addplot[color=tolgreen, densely dashed, mark=square, mark size=1.2pt, forget plot] coordinates {(2024,73.1)(2025,74.9)};
\addplot[color=tolteal, mark=triangle*, mark size=1.5pt] coordinates {(2016,56.1)(2017,50.2)(2018,42.5)(2019,43.4)(2020,36.9)(2021,32.3)(2022,30.0)(2023,28.4)(2024,25.8)};
\addlegendentry{On the merits}
\addplot[color=tolteal, densely dashed, mark=triangle, mark size=1.5pt, forget plot] coordinates {(2024,25.8)(2025,24.1)};
\addplot[color=tololive, mark=*, mark size=1.2pt] coordinates {(2016,7.6)(2017,7.9)(2018,6.5)(2019,6.4)(2020,6.8)(2021,6.3)(2022,5.5)(2023,5.1)(2024,4.4)};
\addlegendentry{Inadmissibility / inoperativeness}
\addplot[color=tololive, densely dashed, mark=o, mark size=1.2pt, forget plot] coordinates {(2024,4.4)(2025,3.4)};
\end{axis}
\end{tikzpicture}}
\caption{Yearly shares, 2016--2025: the cassation side (left) and the rejection side (right). 2025 is incomplete.}
\label{fig:trends}
\end{figure}

Figure~\ref{fig:trends} shows that the cassation rate is stable over time, staying around a third over the decade, while partial cassations gain thirteen points (from $53.6\%$ in 2016 to $66.5\%$ in 2024). Among the grounds, violation of the law rises and lack of legal basis falls (Figure~\ref{fig:trends}, left). Scholarship has measured the share of partial cassations for 2017--2021~\cite{cottin2023cassations}, but the upward trend itself has, to our knowledge, not been reported before. The composition of the rejections moves even more (Figure~\ref{fig:trends}, right). Summary rejection carried $46.4\%$ of the rejected grounds of 2016 and $73.1\%$ of those of 2024, a rise of almost twenty-seven points, while rejection on the merits and inadmissibility both decline. The caseload does not explain this rise. The number of rulings falls over the same period, from 9,685 in 2016 to 8,074 in 2024. The Court made explicit in 2020 that it reserves reasoned rejections for cases with normative value~\cite{ferrand:hal-02943511}: for the other grounds, the case law only records that they failed. A ground-level count like ours makes visible the extent of that silence, how many grounds it covers and how fast it spreads.

\paragraph{Which texts the Court mobilizes\label{sec:res_articles}}

Of all pairs, $74.2\%$ cite at least one article in their reasons that we can resolve to a code. The summary-rejection blocks almost always cite article 1014 CPC (\S\ref{sec:res_reject}). The article layer (\S\ref{sec:articles}) records, for each text, how often it is cited, accepted, and put in the visa. Table~\ref{tab:articles} shows three contrasted examples: a substantive text mainly accepted and put in the visa, a contested one, and a procedural lever that is massively cited but almost absent from the visa. The labeled corpus can be browsed through our interactive web application
. It shows, for each article, how many blocks of reasons cite it, how their outcomes split, and into which doctrinal families.

\begin{table}[t]
\centering
\caption{Three article profiles from the article layer.}
\label{tab:articles}
\scriptsize
\setlength{\tabcolsep}{4pt}
\begin{tabular}{@{}lrrr@{}}
\toprule
Text & Citations & Accepted & In visa \\
\midrule
Art.\,1134 Civil Code, former numbering (binding force of contracts) & $1{,}576$ & $90.1\%$ & $90.2\%$ \\
Art.\,L.\,1154-1 Labour Code (proof of workplace harassment)         & $324$     & $46.0\%$ & $46.9\%$ \\
Art.\,L.\,411-3 COJ (cassation without referral)   & $2{,}281$ & $95.2\%$ & $3.1\%$  \\
\bottomrule
\end{tabular}
\end{table}


\section{Limitations and Conclusion}\label{sec:discussion}

\paragraph{Limitations and perspectives} The main limitation comes from the rules: the doctrinal families are read from canonical wording, so that unusual phrasing escapes the grids
. The specific family counts are therefore lower bounds. Violation of the law, last in the cascade (\S\ref{sec:grounds}), may absorb unusually worded cases of the others. The taxonomies themselves, like any doctrinal systematization, can be discussed~\cite{boucard2017jcl}. We follow Boré's~(\S\ref{sec:bg_cassation}). Further constraints come from the source. Summary rejections, by nature, do not say why a ground failed. Judilibre zoning, extensive only from 2016~(\S\ref{para:corpus:source}), sometimes merges several grounds into one~(\S\ref{sec:valresults}). LLM-based segmentation could lift this dependence.

\paragraph{Conclusion} The dataset turns the case law of the five civil chambers into a queryable, ground-level resource: 121,536 pairs labeled with their outcomes and 117,073 of them attached to their doctrinal families. 
Two families make up $80\%$ of the accepted grounds, summary rejection keeps rising, and cassations become steadily more partial. The grids are specific to French civil law: future work includes rewriting them for the criminal chamber. The Court's own open-data report calls for systematic and exhaustive studies of case law~\cite{cadiet2022diffusion}. We release ours: the dataset, the code and the tool.

\bibliographystyle{vancouver}
\bibliography{references}

\end{document}